\documentclass[review]{elsarticle}

\usepackage{lineno,hyperref}
\usepackage {mathtools}
\usepackage{url}
\usepackage{graphicx}
\usepackage{subfigure}
\usepackage{multirow}
\modulolinenumbers[5]

\journal{Journal of \LaTeX\ Templates}

\biboptions{authoryear}

\begin{document}

\begin{frontmatter}

\title{Aesthetic Quality Assessment for Group photograph}


\author[mymainaddress,mysecondaryaddress]{Yaoting Wang}
\ead{1831125497@tiangong.edu.cn}
\author[mymainaddress,mysecondaryaddress]{Yongzhen Ke\corref{mycorrespondingauthor}}
\cortext[mycorrespondingauthor]{Corresponding author}
\ead{keyongzhen@tiangong.edu.cn}
\author[mymainaddress,mysecondaryaddress]{Kai Wang}
\ead{gllw915@163.com}
\author[mymainaddress,mysecondaryaddress]{Cuijiao Zhang}
\ead{1831125492@tiangong.edu.cn}
\author[mythirdaddress]{Fan Qin}
\ead{qinfan@nankai.edu.cn}

\address[mymainaddress]{School of computer science and technology, Tiangong University, Tianjin, 300387, China}
\address[mysecondaryaddress]{Tianjin Key Laboratory of Autonomous Intelligence Technology and Systems, Tianjin,300387, China}
\address[mythirdaddress]{Business School, Nankai University, Tianjin, 300072, China}

\begin{abstract}
Image aesthetic quality assessment has got much attention in recent years, but not many works have been done on a specific genre of photos: Group photograph. In this work, we designed a set of high-level features based on the experience and principles of group photography: Opened-eye, Gaze, Smile, Occluded faces, Face Orientation, Facial blur, Character center. Then we combined them and 83 generic aesthetic features to build two aesthetic assessment models. We also constructed a large dataset of group photographs - GPD- annotated with the aesthetic score. The experimental result shows that our features perform well for categorizing professional photos and snapshots and predicting the distinction of multiple group photographs of diverse human states under the same scene.
\end{abstract}


\end{frontmatter}

\section{Introduction}
\label{}
With the rapid growth of image applications, the traditional image quality evaluation no longer satisfies the practical need. Thus the image aesthetic quality assessment (IAQA) was born. IAQA uses the computer simulation of human perception and cognition of beauty to automatically assess the "beauty" of images (i.e., Computer evaluation of image aesthetic quality) \citep{1}. It mainly responds to the aesthetic stimuli formed by the image under the influence of aesthetic elements such as composition, color, luminance, and depth of field.

In daily life, we often encounter a situation where we need to take group photographs for souvenirs. So how to estimate the aesthetic values of a group photograph and further provide a guidance system for real-time group photo shooting will become meaningful. The current methods of IAQA mainly focus on the effects of composition, color, light and shadow, depth of field, and other components on the aesthetics of the entire image, which can classify the professional photographs and snapshots, as shown in Figure1(a). Nevertheless, when evaluating the aesthetics of group photographs, people are not only concerned with the above factors but also focus on the state of the person in the image, such as whether someone's eyes closed, does not look at the camera, the face is blocked, does not smile and other factors. If these factors are not taken into account in the aesthetic quality assessment of group photography, the evaluation will not be accurate. An example is shown in Figure 1(b), the two images assessed by the general method have similar ratings. However, when considering the criteria for group photography, it is clear that we are more satisfied with the first one.
\begin{figure}[htp]  
	\centering\includegraphics[scale=0.9,trim=0 0 0 0]{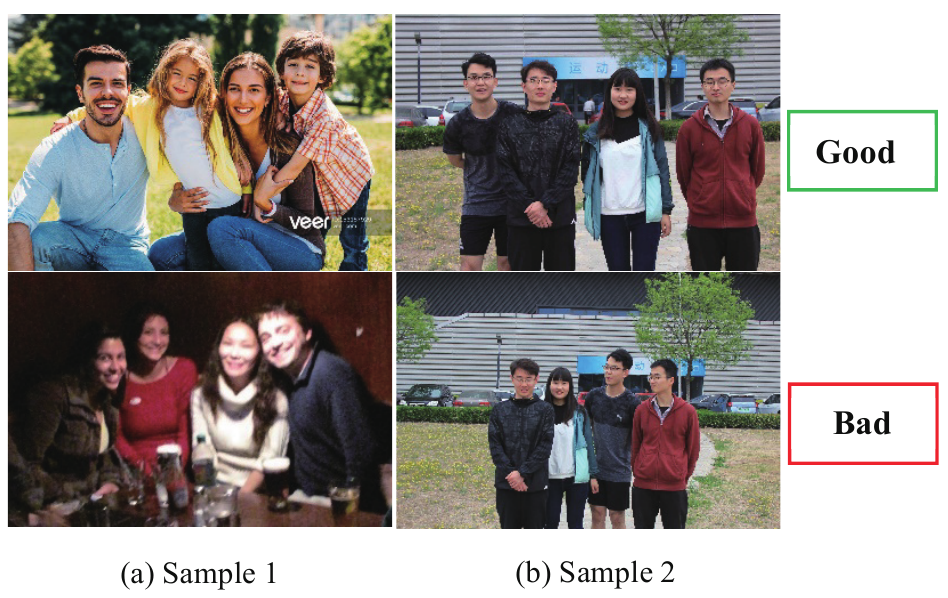}
	\caption{(a) The professional photographs and snapshots. (b) The group photographs of diverse human states under the same scene}
\end{figure}

Therefore, we propose a method to assess the aesthetic quality of the group photograph. Firstly, we extracted the texture, brightness, low depth of field, color, and other features commonly used in IAQA. Moreover, we designed seven specific features that conform to group photography experience and principles, such as whether the face in the photo is occluded, whether someone closes their eyes, whether they smile or not. We then constructed a dataset (GPD) specifically for group photography aesthetic quality assessment. It contains a total of 1000 pictures, which are selected from the network, the existing IAQA dataset, and photos taken by ourselves. Finally, with the extracted features, we trained a classifier and a regression model on the GPD-dataset, which are used to classify photos into good and bad categories and predict aesthetic scores, respectively.

To summarize, our main contributions in this paper are as follows:
\begin{itemize}
	\item We design 7 new features related to group photography as the standard for aesthetic assessment.
	\item We build GPD dataset annotated with the aesthetic score, and develop an online annotation system to collect users' aesthetic evaluation of photographs.
	\item We propose a classifier and a regression model trained by our features, which outperform existing methods in in group photo aesthetic evaluation.
\end{itemize}

\section{Related works}
Early research on IAQA focused on low-level visual features and then training classifiers or regressors to evaluate image aesthetics \citep{2,3,4,5}. In 2004, Microsoft Asia Research Institute and Tsinghua University \citep{5} jointly proposed a method that can automatically distinguish the photographs taken by professional photographers from those taken by customers. This work is considered as the earliest research on IAQA. They used a 21-class, a total of 846-dimensional low-level global features to learn the classification model to classify the test images aesthetically. In 2006, \cite{3} began to use local features for aesthetic assessment, combining the low-level features such as color, texture, shape, picture size and high-level features such as depth of field, tripartite rule, regional contrast and so on which are usually used for image retrieval, and then trained the SVM classifier for the binary classification of image aesthetic quality. \cite{4} proposed using global edge distribution, color distribution, hue counting, contrast, and brightness to represent the image. Based on these features, the naive Bayesian classifier is trained. All the above works are aesthetic evaluations that are unrelated to content. Since 2010, some aesthetic assessment research related to content has appeared \citep{6,7,8}. In 2014, with the emergence of AVA \citep{9}, a large-scale aesthetic analysis dataset, significant progress had been made in the automatic aesthetic analysis by using deep learning technology \citep{10,11,12,13,14}. The classification accuracy rate of ILGNet-Inc.V4 proposed by \citep{10,11,12} ranked first in the world on the ava dataset. In recent years, researchers have mainly studied the problem of IAQA from different tasks. In order to solve the problem of the need for subjective labeling when the image database is established, Ning Ma \cite{15} proposed a deep attractiveness rank net (DARN) model to learn aesthetic scores. \cite{16} proposed a query-based aesthetic assessment deep learning model that makes different aesthetic evaluations based on different styles of images. \cite{17} considers not only objective factors but also subjective factors of user reviews for aesthetic assessment. \cite{18} extended the one-dimensional score to a multi-dimensional aesthetic space score. \cite{19} proposed an A-Lamp CNN architecture to learn the fine-grained and the overall layout aesthetic assessment simultaneously. In addition, in terms of the prediction of image aesthetic distribution, jinxing et al. \cite{20} proposed the method that predicts image aesthetic distribution, opening the direction of aesthetic prediction in the era of deep learning. There are also some research results in the aesthetic assessment of faces and portraits \citep{21,22,23}. In the study of group photo images, \cite{24} proposed the spring-electric model, which recommended the appropriate station and the proportion of the characters to the photograph. However, they did not evaluate the aesthetic quality from the perspective of the subject. To our best knowledge, there are currently no research on the aesthetic assessment of group photography.

\section{Aesthetic Factors for Group photography}
In this section, we will discuss the extraction of features for representing the aesthetic quality of a group photograph. We extract two major groups of features: group photographic features conformed to the group photographic rules and low-level generic aesthetic features proposed in \citep{3,27,28}. These features are combined to obtain better estimates of the aesthetic scores. The following subsections explain each group of features.
\subsection{Group Photography Features}
When people assess group photographs, they usually pay more attention to the facial information and position of the person in the image. Therefore, we utilize proven face recognition tools \cite{face} and \cite{baidu} to extract face-related information and perform further feature design based on this information.

Assuming $N$ faces are detected in a group photograph, the detected face sequence $F$ is represented as follows:
\begin{equation}
	F=\{face^1,face^2,\dots,face^i\} \quad i\in\{1,2,\dots,N\},
\end{equation}
where $face^i$ represents the facial information of the ith person, which includes: The coordinates of the top-left point of the facial box ($x,y$); The height and width of the facial box ($h,w$); Confidence ($c_i$) of different eye states ($S$ broken into 6 states, detailed in section 3.1.1); The gaze direction of the left and right eyes ($D_l,D_r$); The value of the smile ($m$); Rotation angle of the head ($\gamma$); The occlusion degree of seven regions of the face ($o_i$,detailed in section 3.1.2); The position coordinates of the person in the photograph ($P$); The degree of blur of the face ($b$).
\begin{figure}[htp]  
	\centering\includegraphics[scale=0.9,trim=0 0 0 0]{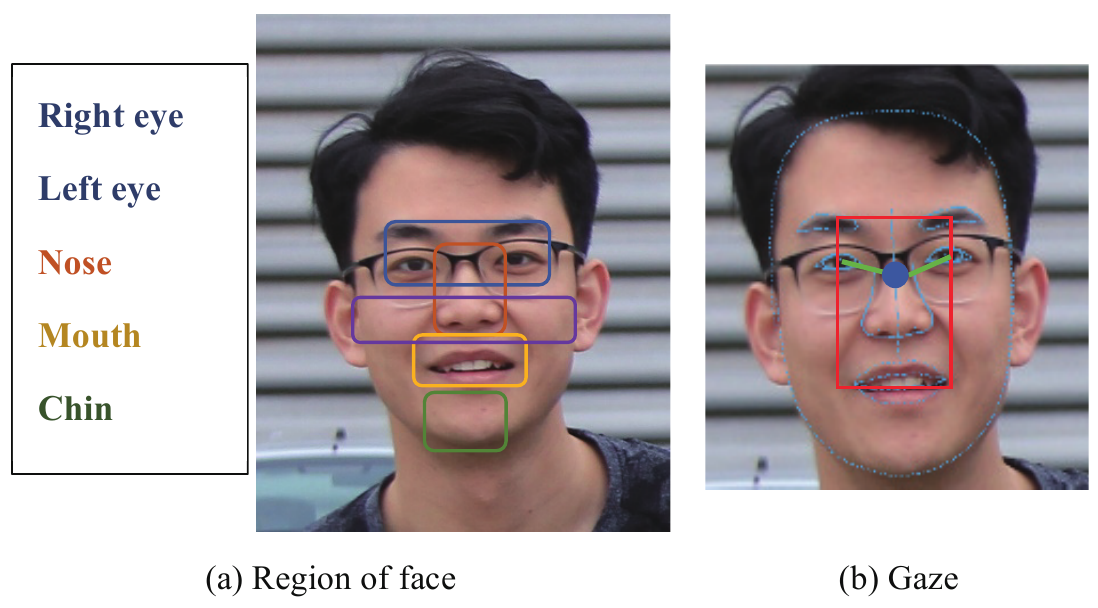}
	\caption{(a) Seven regions of face. (b) Red: range of looking the lens, green: direction of gaze, blue: junction of the gaze.}
\end{figure}
\subsubsection{Open-eyed}
Eyes are the windows to one's soul. When we shoot portraits, we tend to focus on people's eyes, and so do group photos. If someone's eyes are closed or obscured in a group picture, the beauty of that will be greatly decreased. Therefore, we consider the eyes states of each person in the photo, which including: Eye opening without glasses $S_1$; Eye opening with ordinary glasses $S_2$ ; Wearing sunglasses $S_3$; Covered eye $S_4$; Eye closing without glasses $S_5$ and eye closing with ordinary glasses $S_6$. We use \cite{face} to predict the confidence of each status of the left and right eye, respectively, which are \{$c_1,c_2,c_3,c_4,c_5,c_6$\}. The $\sum_{i=1}^{6} c_i$ equals 100. We select the state of the maximum of six confidences as the condition of the eye. If the condition of eyes belongs to one of these three states ($S_1,S_2,S_3$), we judge the person is open-eyed and thus formulate that as:
\begin{equation}
S_r=f(\max_{1\leq j\leq 6}c_j^{(r)}),
\end{equation}
\begin{equation}
S_l=f(\max_{1\leq j\leq 6}c_j^{(l)}),
\end{equation}
\begin{equation}
E_i=\begin{cases}
1  &  \text{if} \; S_l \in \{S_1,S_2,S_3\} \text\;{and} \;S_r \in \{S_1,S_2,S_3\} \\
0  &  \text{otherwise}
\end{cases}
\quad  i\in\{1,2,\dots,N\},
\end{equation}
where $S_l$ and $S_r$ represent the final state prediction of the left and right eyes, respectively. $c_j^{(l)}$ and $c_j^{(r)}$  are the confidence of 6 states corresponding to the left and right eyes. $f$ is the mapping between confidence and corresponding status. We further calculate the proportion of people whose open-eyed. Based on the experience for group photograph assessment, we found the proportion has a non-linearity relationship with the evaluation result. i.e., when all the people open their eyes, the assessment is high. Once someone closes their eyes, the evaluation of the image falls into the bad category, and then gradually decrease with the number of close-eyed people. Thus we fit the formula (5), where  $\frac{1}{N}\sum_{i=1}^{N} E_i$ is the proportion and $f_1$ is the final feature of open-eyed.
\begin{equation}
f_1=\begin{cases}
1  &  \text{if}  \;\frac{1}{N}\sum_{i=1}^{N} E_i=1 \\
1-2^{-\frac{1}{N}\sum_{i=1}^{N} E_i}  &  \text{otherwise}
\end{cases}
\quad  i\in\{1,2,\dots,N\},
\end{equation}

\subsubsection{Occluded Faces}
In group photography, the most basic requirement is that everyone's face is not occluded. If someone is masked, no matter how splendid the color, composition, light and shadow of the photo is, the photo will be discarded without hesitation. Hence, whether the face is occluded or not is another crucial criterion for judging the quality of the group photograph.

We use the method provided by \cite{baidu} to obtain the information about occluded faces in the photograph. The face is segmented into seven regions (see Figure 2. (a)): the left and right eye, the left and right cheek, mouth, jaw, and nose. An occlusion degree will be calculated of each region, which is a floating-point number in the range [0, 1], where 1 means that the region is completely occluded. When the occlusion degree of any region exceeds the recommendation threshold provided in BaiduAI, we judge that the person's face is occluded. We define the occluded face as:
\begin{equation}
O_i=\begin{cases}
1 &  \exists\; o_i \geq \theta_j \\
0 &  \text{otherwise}
\end{cases}
\quad  i\in\{1,2,\dots,N\},j\in\{1,2,\dots,7\},
\end{equation}
where $O_i$ indicates whether the face of the ith person is occluded, and 1 denotes occlusion. The $o_j$ and $\theta_j$ are the value of the occlusion degree and threshold of each region.
Then we calculate the proportion of the number of un-occluded people defined as ($1-\frac{1}{N}\sum_{i=1}^{N} O_i$). Same as $f_1$, the proportion and evaluation also satisfies the nonlinearity relationship. The $f_2$ is the occluded faces feature, and the formula is described as follows.
\begin{equation}
f_2=\begin{cases}
1  &  \text{if}  \;\sum_{i=1}^{N} O_i=0 \\
1-2^{-(1-\frac{1}{N}\sum_{i=1}^{N} O_i)}  &  \text{otherwise}
\end{cases}
\quad  i\in\{1,2,\dots,N\}.
\end{equation}
\subsubsection{Face Orientation}
A word that photographers often say during photography is "Looking at the camera". If a person in viewfinder looks at the camera, but the head tilted, the photo is not a high-quality image. Therefore, we get the yaw angle of head as $\gamma$, where  $\gamma$ $\in$ [-180,180]. When $\gamma$ $\in$ [-30,30], it is considered that facing the camera. We record whether the character is facing the camera as $H_i$, which equals  means yes. Same as above, the proportion of the number of people without head-tilted to the total number of people $N$ is calculated. This proportion also fits a non-linear relationship. $f_3$ is the face orientation feature. The formula is as follows.
\begin{equation}
H_i=\begin{cases}
1 &  \gamma_i \in[-30,30]\\
0 &  \text{otherwise}
\end{cases}
\quad  i\in\{1,2,\dots,N\},
\end{equation}
\begin{equation}
f_3=\begin{cases}
1  &  \text{if}  \;\frac{1}{N}\sum_{i=1}^{N} H_i=1 \\
1-2^{-\frac{1}{N}\sum_{i=1}^{N} H_i}  &  \text{otherwise}
\end{cases}
\quad  i\in\{1,2,\dots,N\},
\end{equation}
\subsubsection{Gaze}
For formal group photography, everyone's focus on the lens is an important criterion. So we designed a feature to represent the proportion of people looking at the camera. There are three prerequisites before estimating gaze: eyes open, facing the camera, eyes are not occlusion. We utilize the information detected by face++ to calculate the direction of the gaze. The gaze estimation process is as follows:
\begin{enumerate}[1.]
	\item Determine the center of the circle: $O=(C_1+C_2)/2$.
	\item Determine the radius: $R=max(w,h)$.
	\item Calculating the average gaze: $D=(D_l+D_r)/2$.
	\item Calculating the gaze junction point coordinates: $p=O+R*D$.
\end{enumerate}
Where $C_1$ and $C_2$ are the landmarks of the left and right eyeball center, $w$ and $h$ are the width and height of the rectangle of face, respectively. The gaze direction vector of the left and right eye are recorded as $D_r=(x_r,y_r)$ and $D_l=(x_l,y_l)$. We use the face landmarks to define a rectangular range (see Figure 2.(b)), If the gaze junction point falls within the range, it is judged that the people is looking at the lens. It is defined as follow,
\begin{equation}
G_i=\begin{cases}
1 &  p_i \in Range_i\\
0 &  \text{otherwise}
\end{cases}
\quad  i\in\{1,2,\dots,N\},
\end{equation}
where $P_i$ represents the coordinates of the gaze junction of the ith person in the frame, and $Range_i$ is the rectangular range of the ith person,  $G_i=1$ means looking at the camera. Then we take the ratio of the people looking at the lens, and the ratio also meets the nonlinear relationship with the assessment. $f_4$ is the gaze feature.
\begin{equation}
f_4=\begin{cases}
1  &  \text{if}  \;\frac{1}{N}\sum_{i=1}^{N} G_i=1 \\
1-2^{-\frac{1}{N}\sum_{i=1}^{N} G_i}  &  \text{otherwise}
\end{cases}
\quad  i\in\{1,2,\dots,N\},
\end{equation}
\subsubsection{Facial blur}
Whether the face is clear or not is essential to the quality of a photograph. Therefore, we obtain the blur degree $b_i$ of face by Face++. We employ the recommended threshold  $v$ (generally $v$ is 50) as the threshold. If the blur degree greater than the threshold, we consider the person’s face was not captured clearly. It can be formalization as:
\begin{equation}
B_i=\begin{cases}
1 &  b_i > v \\
0 &  b_i \leq v
\end{cases}
\quad  i\in\{1,2,\dots,N\},
\end{equation}
where $B_i$ indicates whether the face of the ith person in the photograph is blurred or not. Then we calculated the percentage of the number of people whose facial blur degree exceeded the threshold as $\frac{1}{N}\sum_{i=1}^{N} B_i$. The higher the percentage, the higher the quality. $f_5$ is facial blur feature.
\begin{equation}
f_5=\begin{cases}
1  &  \text{if}  \;\sum_{i=1}^{N} B_i=0 \\
1-\frac{1}{N}\sum_{i=1}^{N} B_i  &  \text{otherwise}
\end{cases}
\quad  i\in\{1,2,\dots,N\}.
\end{equation}

\subsubsection{Smile}
Smile plays a vital role in the emotional expression of group photography. Through observation, we found that a large proportion of people smiling in group photograph is often attractive and easier to remember than no smile in group photograph. We use $m$ to represent the degree of smile. There is a threshold $w$ for the degree of smile provided by Face++.  We count the number of people with a smile formulated as $\sum_{i=1}^{N} M_i$, which the degree of smile greater than the threshold. Then we take the ratio of the number of people with smile as the smile feature $f_6$, which is defined as:
\begin{equation}
M_i=\begin{cases}
1 &  m_i > w \\
0 &  m_i \leq w
\end{cases}
\quad  i\in\{1,2,\dots,N\},
\end{equation}
\begin{equation}
f_6=\begin{cases}
1  &  \text{if}  \;\sum_{i=1}^{N} M_i=0 \\
\frac{1}{N}\sum_{i=1}^{N} M_i  &  \text{otherwise}
\end{cases}
\quad  i\in\{1,2,\dots,N\}.
\end{equation}

\subsubsection{Character center}
Through observation and experience, we found that in a good group photo, the positions of the people are usually horizontally centered and uniformly arranged, particularly the formal group photos. Therefore, the horizontal position of people is also positively correlated to the quality of group photograph.We sequentially detect the horizontal x-axis coordinate of the center of each person's face, represented as $x_i$, and then average the x-axis coordinates, which is represented by $P_x$ define as:
\begin{equation}
P_x=\frac{\sum_{i=1}^{N}x_i}{N}\quad  i\in\{1,2,\dots,N\}.
\end{equation}
Next, we compute the relative position $R$ of the character center and the picture. We formulate that as $R=P_x/W$, where $W$ is the width of the frame. We divided the photograph evenly into five parts. If $R$ is in the range of 0.4 to 0.6, it means that $R$ is located in the center of the photograph, i.e., the people position is horizontally centered. We call $f_7$ the character center feature.
\begin{equation}
f_7=\begin{cases}
1  &  0.4\leq R\leq 0.6 \\
0  &  \text{otherwise}
\end{cases}
\quad  i\in\{1,2,\dots,N\}.
\end{equation}

\subsection{Generic Aesthetic Features}
In addition to group photographic features, we selected 83 features from the generic aesthetic features mentioned in \citep{3,27,28}, such as exposure, saturation and texture based on wavelet transform, as aesthetic features to evaluate group photography aesthetics. These features can be divided into four categories: color, local, texture, and composition. The above features are not the focus of this paper, so briefly described in Table 1.

\begin{table}[tb]
	\scriptsize
	\begin{tabular}{p{1.5cm}|p{2cm}|c|p{6.5cm}}
		\hline
		\textbf{Category}& \textbf{Short Name}& \textbf{\#} & \textbf{Description} \\ \hline
		\hline
		\multirow{4}{*}{color} & Brightness, Hue, Saturation & f8-f13 & mean brightness, saturation, and hue of the image and the center of the picture \citep{3}. \\ \cline{2-4} 
		                                  & Emotion                 & f52-f54         &emotional measure based on brightness and saturation\citep{28}.                                     \\ \cline{2-4} 
		                                  & Colorfulness              & f55             & colorfulness measured, using the Earth Mover’s Distance (EMD) between the histogram of an image and the histogram having a uniform color distribution \citep{28}.        \\ \cline{2-4} 
		                                  & Color                     & f56-f71         & amount of black, silver, gray, white, maroon, red, purple, fuchsia, green, lime, olive, yellow, navy, water blue \citep{27}.  \\ \hline
		\multirow{3}{*}{regional} & Disconnected Region        & f28             & image segmentation, based on K-means, number of disconnected regions in the image \cite{3}.    \\ \cline{2-4} 
		                                   & Local HSV                  & f29-f43         & average H, S and V values for each of the top 5 connected regions \citep{3}.                                                                                            \\ \cline{2-4} 
		    								& Ratio                     & f44-f48         & the size ratio of the top 5 connected regions with respect to the image. \citep{3}                                                                                       \\ \hline
		\multirow{2}{*}{texture} & Wavelet textures            & f14-f25         & after three-level wavelet transform, wavelet textures for each channel (Hue, Saturation, Brightness) and each level (1-3), sum of all levels for each channel \citep{3}.  \\ \cline{2-4} 
											 & GLCM              & f72-f83         & features based on the GLCM: contrast, correlation, homogeneity, energy for Hue, Saturation and Brightness channel \citep{28}.                                            \\ \hline
		\multirow{4}{*}{composition} & Image size                 & f26-f27         & Image size, sum of the length and width; image proportion, ratio of the length and width. \cite{3}                                                                      \\ \cline{2-4} 
		                                     & Low Depth of Field (DOF)    & f49-f51         & low depth of field indicator; ratio of wavelet coefficients of inner rectangle vs. whole image (for Hue, Saturation and Brightness channel) \citep{3}.                   \\ \cline{2-4} 
		                                     & Dynamics                 & f84-f89         & absolute angles, relative angles, and lengths of static (horizontal, vertical) and dynamic (oblique) lines \citep{28}.                                                   \\ \cline{2-4} 
		                                     & Level of Detail         & f90             & number of segments after waterfall segmentation \citep{28}.  \\ \hline
	\end{tabular}
\caption{83 generic aesthetic features}
\end{table}
\section{Group photography Dataset}
The datasets related to photography aesthetics include AVA \citep{9}, AADB \citep{13}. AVA included 250,000 images, each with the corresponding aesthetic classification and rating labels. AADB contains 10,000 images, which has more balanced distribution of professional photograph and snapshot. Each image is annotated with score and eleven attributes. Nevertheless, there is no dataset for aesthetic evaluation of group photography at present. To this end, we collected a group photography dataset GPD by ourselves, which consists of three parts: group photographs shot by ourselves, selected from the existing aesthetic photography dataset, and obtained through internet. GPD contains a total of 1000 group photographs, and each image has been scored. Samples of GPD is shown in Figure 3.

\begin{figure}[htp]  
	\centering\includegraphics[scale=0.85,trim=0 0 0 0]{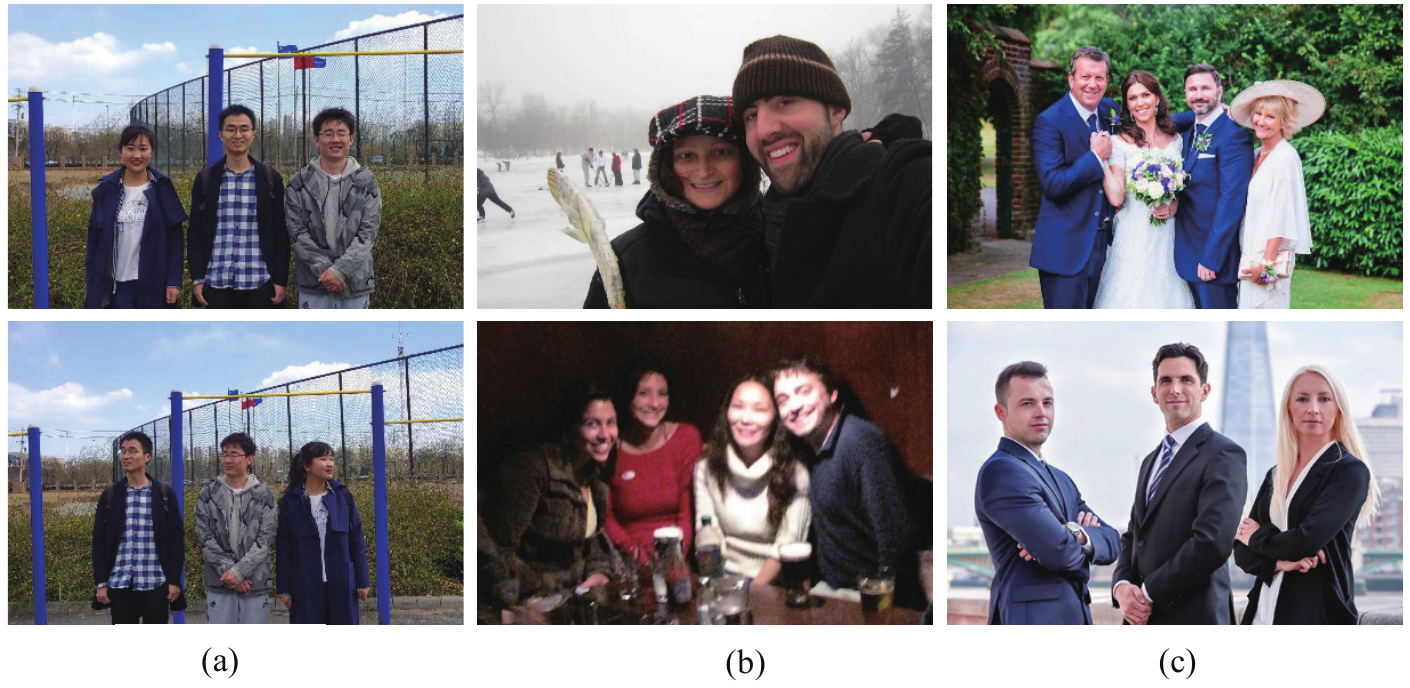}
	\caption{Samples of GPD, (a) taken by ourselves (b) from the existing dataset   (c) from the internet}
\end{figure}

\paragraph{(a)}Shooting by ourselves: Our research team used mobile phones and SLR cameras to take some group photographs. During the photography, the subjects constantly changed position and their expression. Most of the time, the photographer was in the status of continuous shooting, and deliberately took some photographs under the condition of out of focus, overexposure, not following the composition, and blurring caused by shaking the hands. The photographs taken by ourselves are mostly image pairs, i.e., multiple photos of different states are taken in the same scene. This is for better explain the inaccuracy of traditional method evaluating the group photographs. This section contains a total of 600 images.
\paragraph{(b)}Selected from the existing dataset: We selected part of the group photographs from the AVA and AADB datasets. The sources of these images are mostly social networking sites such as Flickr and DPChallenge. Most of the photographs were shot and uploaded by amateur photographers. We selected the group photographs among them, but the aesthetic quality of these is not high, and there are photographic problems such as blurring and overexposure. So this partition balances the distribution of quality in GPD, making GPD more robust. This section contains a total of 224 images.
\paragraph{(c)}Download from the Internet: We selected group photos from image sites such as Baidu Pictures, Petal.com. This partition includes 74 images, all of which are formal group photographs. They are taken by professional photographers and have high aesthetic quality. The photographs in this partition are more attractive than the previous two.
\begin{figure}[!ht]  
	\centering\includegraphics[scale=0.8,trim=0 0 0 0]{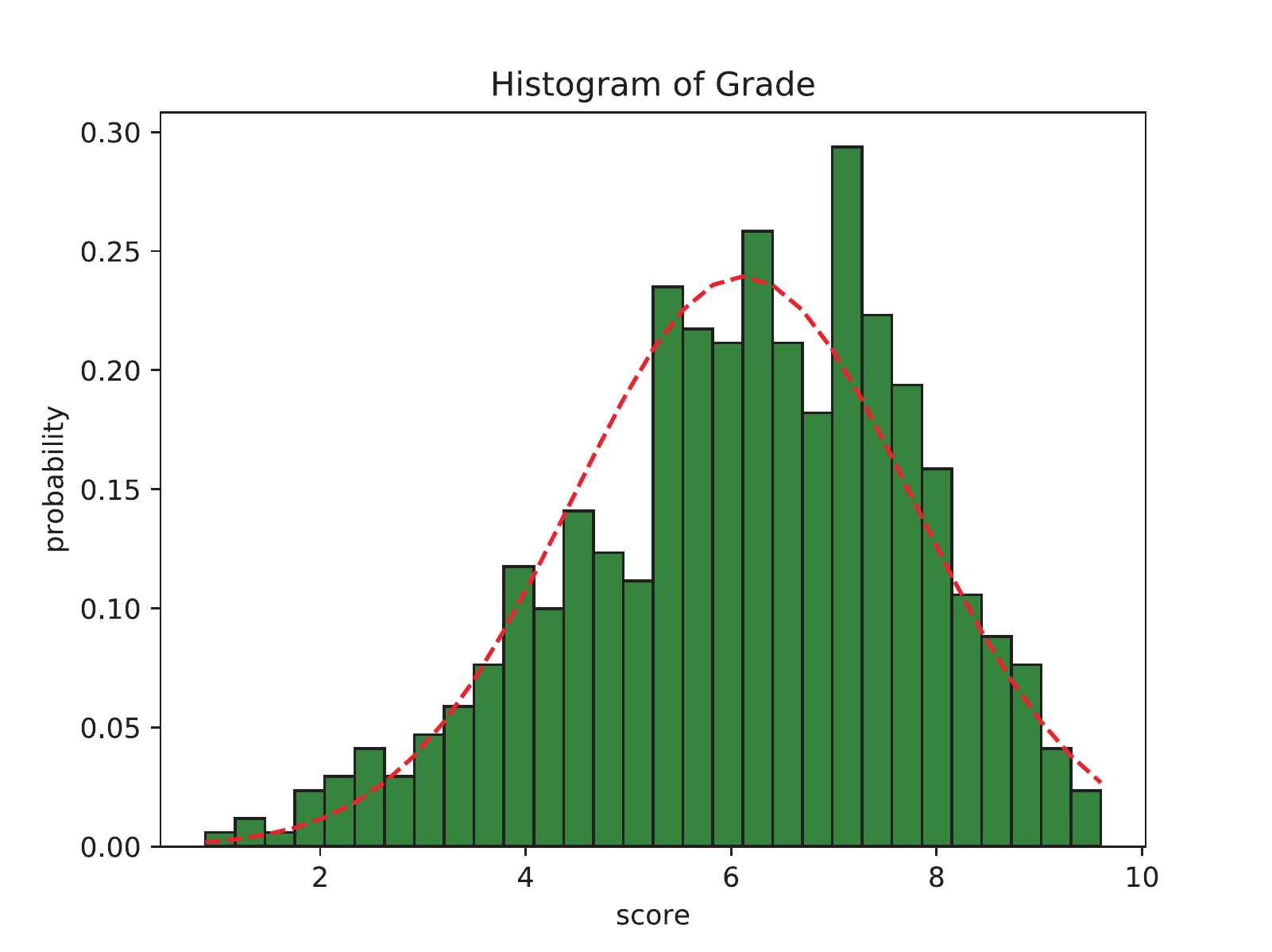}
	\caption{The distribution of aesthetic scores. The horizontal axis represents 0-10 points, and the vertical axis represents the proportion of the corresponding scored pictures to the total number of pictures, conforming the gaussian distribution.}
\end{figure}

To obtain the aesthetic annotations of the group photograph, we designed an online annotation tool, which can rate the group photograph that appears randomly - made the assessment based on the first impression - on the website by users. The scores range from 1 to10. We give tips on the website for scoring, "please pay attention to the following factors when scoring: face occlusion, eyes closed, gaze, smile, and general aesthetic factors such as lighting, composition, color, and picture clarity." In the end, each photograph is assessed by 5 to 20 people, and the average score of each image is taken as its ground truth label. In addition, the website has an image upload model, so users can voluntarily upload their own group photographs for the GPD. Figure 4 shows the probability distribution of GPD.

In GPD dataset, there are two kinds of annotation for each image, one is that the binary value represents the quality of the image, which is used for classifier training, and the other is the score, which is used for regression training. The binary label is obtained by binarizing the score label with 6 (the median value of the aesthetics label in the dataset) as the dividing line

\section{Aesthetic Quality Assessment for Group Photograph}
In order to verify the effectiveness of our proposed group photo aesthetic features, we proposed a method whose system flow as in Figure 5. We first construct a group photo dataset, including the image and the label (ground truth), and then perform image preprocessing on all images. On the processed image, we extract the group photograph features and generic aesthetic features, and store them into a vector. After feature extraction, the dataset is divided into a training and test set to training a classifier and a regression. The classifier classifies the photo into two categories: good or bad. The regressor evaluates the image aesthetics with a score of 1 to 10, and finally uses the trained classifier and regression to predict the photo in the test set. We compared the results with the test set label to estimate the accuracy of the classifier and regressor.
\begin{figure}[htp]  
	\centering\includegraphics[scale=0.8,trim=0 0 0 0]{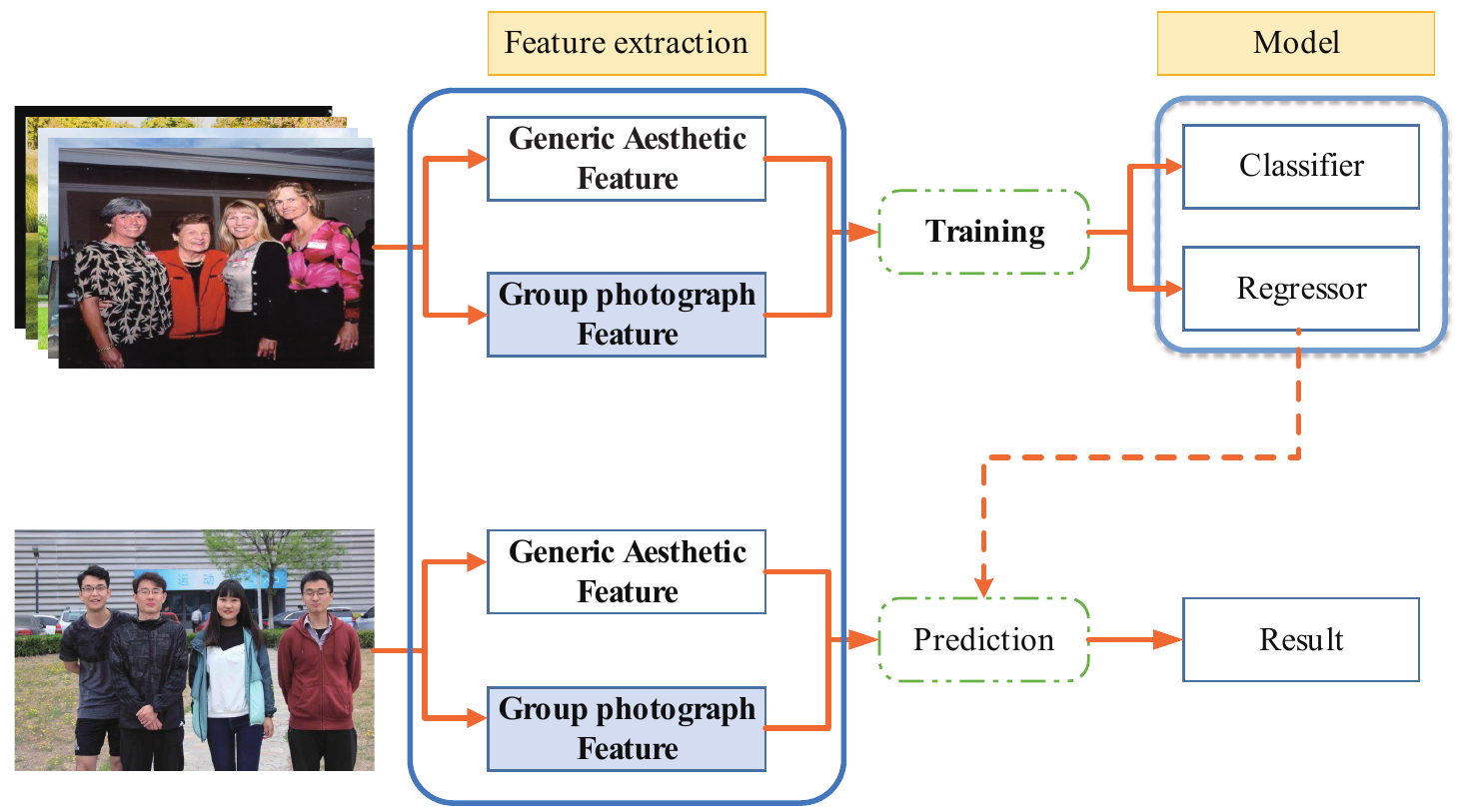}
	\caption{Flow chart of group photo aesthetic evaluation method}
\end{figure}

Before generic aesthetic features extraction, we preprocess all the images. The processes include: Adjusting the image size to 128 *128 pixels, which can not only retain enough image information but also meet the efficiency of calculation; Converting RGB color space into HSV and LUV color space, some features need to be extracted from these two color spaces; The K-means is used to segment the image according to chromaticity in the LUV color space; The Waterfall segmentation \citep{29} is used to segment the image into continuous regions in the HSV color space. On the basis of these image preprocessing, the features are extracted according to the description in Table 1.

Before group photograph feature extraction, we utilized Face++ to detect and save the information of facial recognition, the state of the person's eyes, the smile degree, the rotation angle of face, the degree of facial blur and the landmarks of face in each image from the GPD. We applied Baidu AI's face detection tool to detect and save the face occlusion of the person. Based on these information, the group photograph features are calculated according to Section 3.1.

\section{Experiments}
This section shows the effectiveness of our proposed features, and comparison of the performances of our method with other methods, in the specific genre: group photography. Firstly, we used the random forest to obtain the importance of each feature to analyze their impact on assessment. Secondly, we applied k-fold cross validation (k = 10) to split GPD into train set and test set, then trained a classifier using  support vector machine (SVM) and a regression model using random forest regression (RF). Finally, we report the performance of this method compared with other methods based on deep learning.

\subsection{Importance of features}
Before evaluating the importance of features, all 90 features were normalized by the Z-score standardization method, i.e., using conversion function : $(X-mean)/std$. We used the Gini-based Random Forest \citep{30} to analyze the respective importance ranking of all features for the model. The top 33 features which importance greater than 0.011 are shown in Figure 6.
\begin{figure}[htp]  
	\centering\includegraphics[scale=0.7,trim=0 0 0 0]{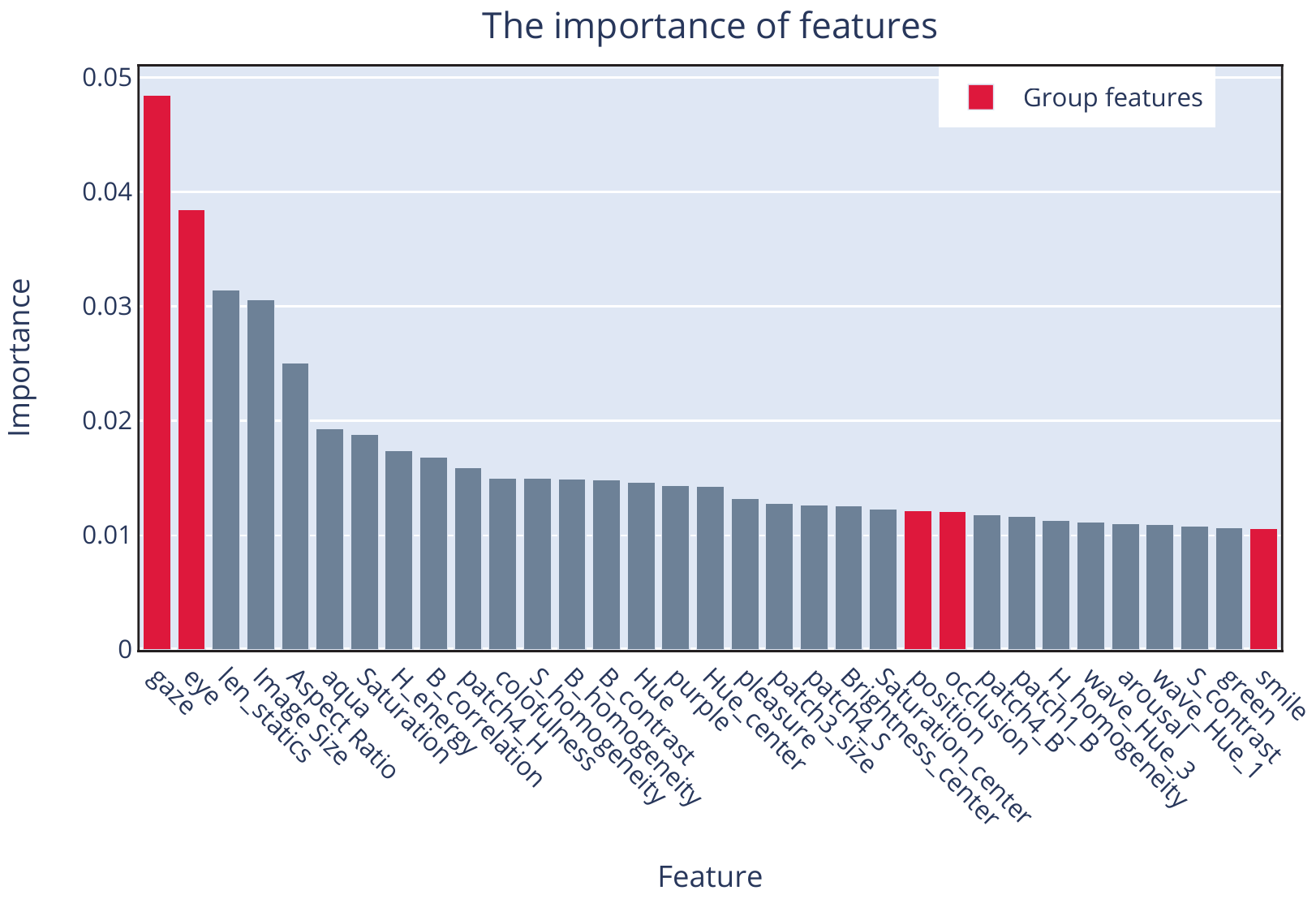}
	\caption{TOP33 feature importance ranking}
\end{figure}

It can be seen that there are 5 group photograph features in the ranks. Among them, the importance of gaze features and opened-eyes features is much higher than other, indicating that the eyes state is important in the group photo evaluation. The importance of the central position of the character, face occlusion, and smile also exceeded the average value, which also played a positive role in the model. The two features of facial blur and face orientation do not appear in top33, because the feature extraction of facial blur depends on the image resolution. If the image itself is low-resolution, the face also blur. The estimation of face orientation is challenging which affected by the light direction, shooting angle, etc., so it is not accurate. The length of the static line is the third important feature, which demonstrates that the feature of horizontal line composition are positive for group photo aesthetic assessment. The three features of brightness, saturation and hue in the center of the image are as same as our hypothesis that the group photography should satisfy the central composition rule. We also found that emotional features (PAD), Pleasure and Arousal has some influence, Pleasure reflects the degree of the people's love for images, The Arousal reflects the level of neurophysiological activation, dominance reflects people's anger and fear, and there is no direct relationship with the evaluation of group photos, which is basically consistent with our hypothesis.
\subsection{Classifier}
Through the feature importance analysis based on random forest, it can be concluded that not all features are effective for group photo assessment. So, as same as \citep{28}, we used two feature selection methods (filter-based and wrapper-based) to filter out the useless features: one is based on the accuracy of single feature classification and the other is recursive feature elimination (RFE) - a feature selection method based on wrapper. We used the sklearn-svm package \citep{sklearn} to train the classification model using the standard RBF kernel ($\gamma = 2.0 .C=1.0$), and use 10-fold cross-validation to ensure the fairness of the experiment. The average AUC of 10-fold cross-validation was adopted as the quality measure of the classifier. AUC is defined as Area under the ROC Curve.

Because the average score of GPD is 6.05, we employed 6 as the boundary to divide the group image into two categories: good and bad. The ROC curve of the model trained by each group photo feature is shown in Figure 7(a), which performance is similar to the importance ranking. The AUC of the gaze feature is 0.73 and the AUC of the opened-eye feature is 0.68. It also shows that the two features are effective for group photo assessment, and the effect of facial blur feature is not ideal, which is due to the challenge of the face recognition in low-quality images.
\begin{figure}[htp] 
	\centering 
	\subfigure[]{ 
		\label{Fig.sub.1} 
		\includegraphics[width=0.48\textwidth]{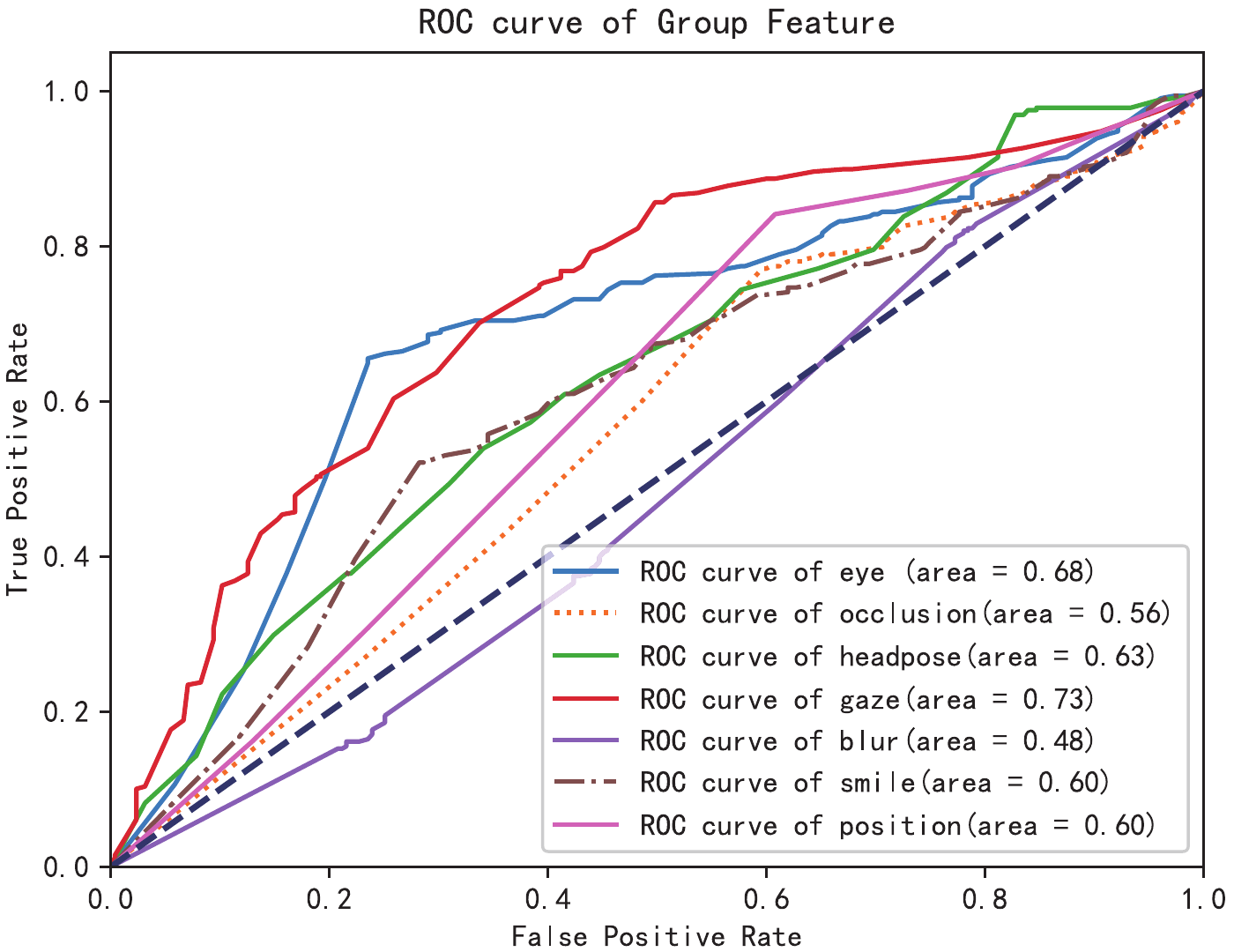}} 
	\subfigure[]{ 
		\label{Fig.sub.2} 
		\includegraphics[width=0.48\textwidth]{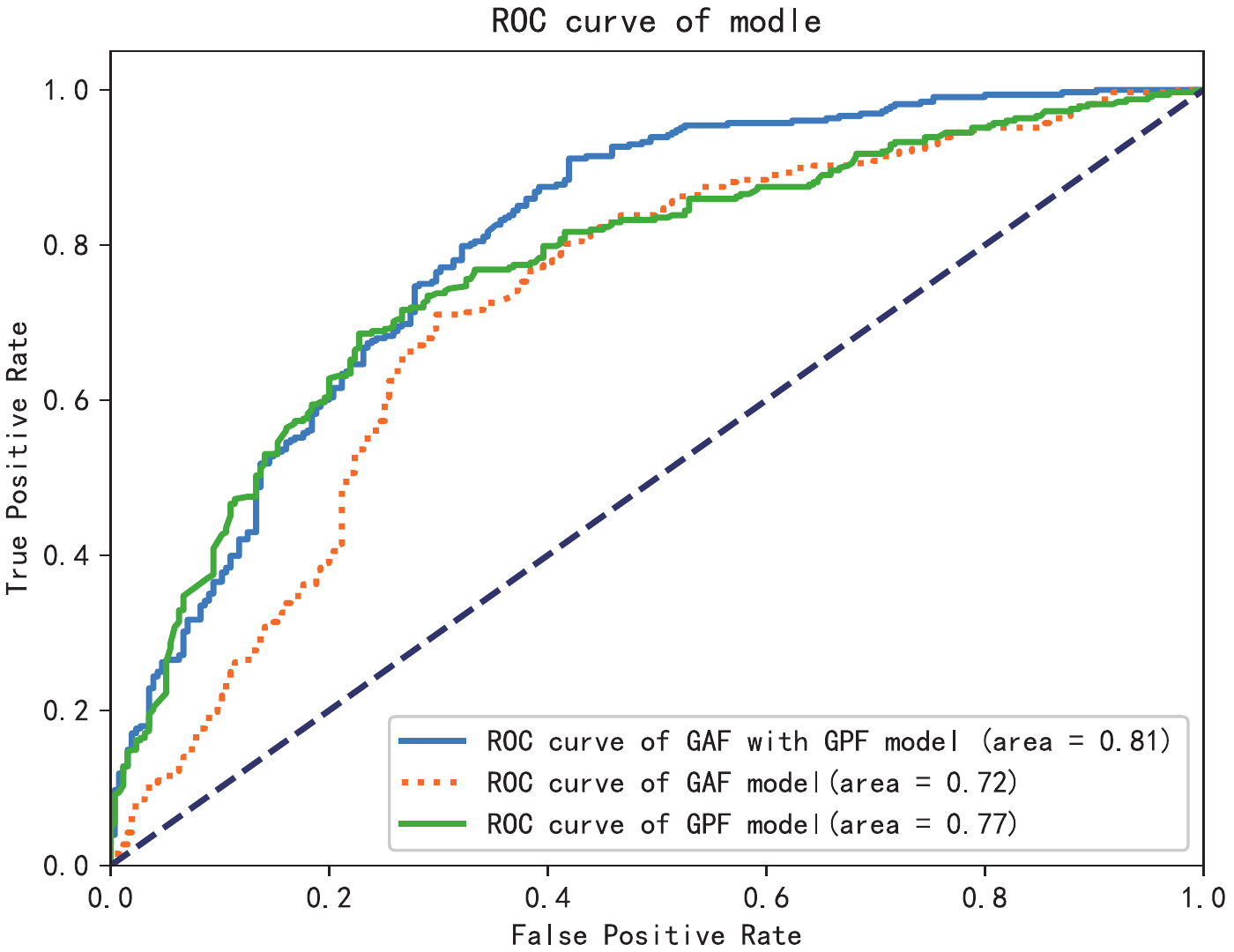}} 
	\caption{(a) ROC of group photo feature model, (b) ROC of three models} 
	\label{Fig.lable} 
\end{figure}

Figure 7 (b) shows the ROC and AUC of three models built on the combination of group photograph features (GPF) and generic aesthetic features (GAF). Among them, the features applied in the GAF\&GPF model are 20 features selected from all features by the above two feature selection methods, the features used in the GAF model are selected from the generic aesthetic features, and the features used in the GPF model only contain group photo features. The selected feature set is shown in Table 2. It can be seen that the GAF\&GPF model completely wraps the GAF model, and the AUC value reached 0.81. The AUC of the GPF model is larger than the GAF model, but smaller than the GAF\&GPF model, which indicates that GAF combined with GPF can make the evaluation performance to the best. Table 3 compares the three models from four measurements: accuracy, precision, recall and F1. We found that the model trained by the combination of the generic aesthetic features and group photo features is better than the other two models in each measurement.
\begin{table}[!h]
\begin{tabular}{|l|l|l|l|l|l|}
	\hline
	\#      & GPF Model & \begin{tabular}[c]{@{}l@{}}GAF\&GPF\\  Classifier\end{tabular} & \begin{tabular}[c]{@{}l@{}}GAF \\ Classifier\end{tabular} & \begin{tabular}[c]{@{}l@{}}GAF\&GPF\\ Regressor\end{tabular} & \begin{tabular}[c]{@{}l@{}}GAF \\ Regressor\end{tabular} \\ \hline
	f1-f7   & *         & *                                                              &                                                           & *                                                            &                                                          \\ \hline
	f8-f13  &           & *                                                              & *                                                         & *                                                            & *                                                        \\ \hline
	f55     &           &                                                                &                                                           & *                                                            & *                                                        \\ \hline
	f56-f71 &           &                                                                &                                                           & *                                                            & *                                                        \\ \hline
	f28     &           & *                                                              &                                                           &                                                              &                                                          \\ \hline
	f29-f43 &           & *                                                              & *                                                         & *                                                            & *                                                        \\ \hline
	f44-f48 &           &                                                                & *                                                         &                                                              & *                                                        \\ \hline
	f14-f25 &           & *                                                              & *                                                         &                                                              &                                                          \\ \hline
	f72-f83 &           & *                                                              & *                                                         & *                                                            & *                                                        \\ \hline
	f26-f27 &           & *                                                              & *                                                         & *                                                            & *                                                        \\ \hline
	f49-f51 &           &                                                                &                                                           &                                                              & *                                                        \\ \hline
	f84-f89 &           &                                                                &                                                           &                                                              & *                                                        \\ \hline
\end{tabular}
\caption{Feature set used by each model, '*' indicates the selected}
\end{table}
\begin{table}[htp]
	\centering
	\begin{tabular}{|l|l|l|l|l|}
		\hline
		Models                & Accuracy & Precision & Recall & F1     \\ \hline
		GAF\& GPF classifier & 0.7097   & 0.7968    & 0.7543 & 0.7285 \\ \hline
		GAF classifier        & 0.6573   & 0.5721    & 0.5969 & 0.5612 \\ \hline
		GPF classifier        & 0.6889   & 0.7878    & 0.6771 & 0.7025 \\ \hline
	\end{tabular}
\caption{Comparison: performance of three models}
\end{table}
\subsection{regression model}
We adopted random forest regression algorithm to train the regression model, through 10-fold cross verification to determine the parameters: the maximum depth is 5, and the number of basic learners is 130. Firstly, the random forest algorithm is used for feature selection. Like the training classifier, three different feature subsets are selected from the feature set which is shown in Table 2. Using these three feature sets to train three models on the GPD (20\% randomly selected as the test set and 80\% as the training set) for 100 times. We use the $R^2$ (coefficient of determination) to measure the regression.  $R^2$ is always between 0 and 1, best score is 1, which is defined as:
\begin{equation}
R^2=1-\frac{\sum_{i}^{N_{test}}(\hat{Y}^{(i)}-Y^{(i)})^2}{\sum_{i}^{N_{test}}(\bar{Y}^{(i)}-Y^{(i)})^2},
\end{equation}
where $\hat{Y}$ is the prediction score, $Y$ is label (ground truth), $\bar{Y}$ is the average value of the test image label, and $N_test$ is the number of test images. Finally, Averaging the $R^2$ of 100 times to avoid the coincidence caused by random sampling from dataset, the comparison of the experimental results is shown in Table 4.
\begin{table}[htp]
	\centering
	\begin{tabular}{|l|l|l|}
		\hline
		Models               & Maximum $R^2$& Average $R^2$\\ \hline
		GAF \& GPF Regressor & 0.563   & 0.415   \\ \hline
		GPF Regressor        & 0.529   & 0.372   \\ \hline
		GAF Regressor        & 0.379   & 0.241   \\ \hline
	\end{tabular}
\caption{Comparison of performance of the three regression models}
\end{table}
The results show that the average $R^2$ of the GAF\&GPF model reaches 0.415 and the Maximum $R^2$ reaches 0.563 in these 100 times trainings, which is not particularly high, but the best performance of the three models. It also shows that the group photo features and general aesthetic features are effective for group photo evaluation. The $R^2$ of GPF model is also higher than that of the model trained by generic aesthetic features, which proved that people pay more attention to the rules we proposed for group photography. General aesthetic features have relatively little impact on the assessment of the group photos.

\subsection{Comparison}
In order to verify that the generic aesthetic features cannot fit the image aesthetic assessment for group photos, and the other method cannot distinguish the photos of different people`s status under the same scene, we preformed the following comparison. We have taken four groups of photos, where each group contains a standard group photo and three photos that do not conform to the group photography rules. They are divided into three categories: "Looking away", "Occlusion" and "Not in the center ". Then we utilized four methods: NIMA-res \citep{11}, NIMA-mobile \citep{11}, Kong \citep{13} and our regression model to evaluate them. The discrimination of the standard image and the other types is defined as $\delta=s_{aes}(I_{standard})-s_{aes}(I_{other})$, where $s_{aes}(I_{standard})$ and $s_{aes}(I_{other})$ are the score of standard group photo and other  types. We calculate the difference between other types and standard photos in each group to measure the discrimination of each model. Figure 8 shows the comparative experimental results.
\begin{figure}[t]  
	\centering\includegraphics[scale=0.83,trim=0 0 0 0]{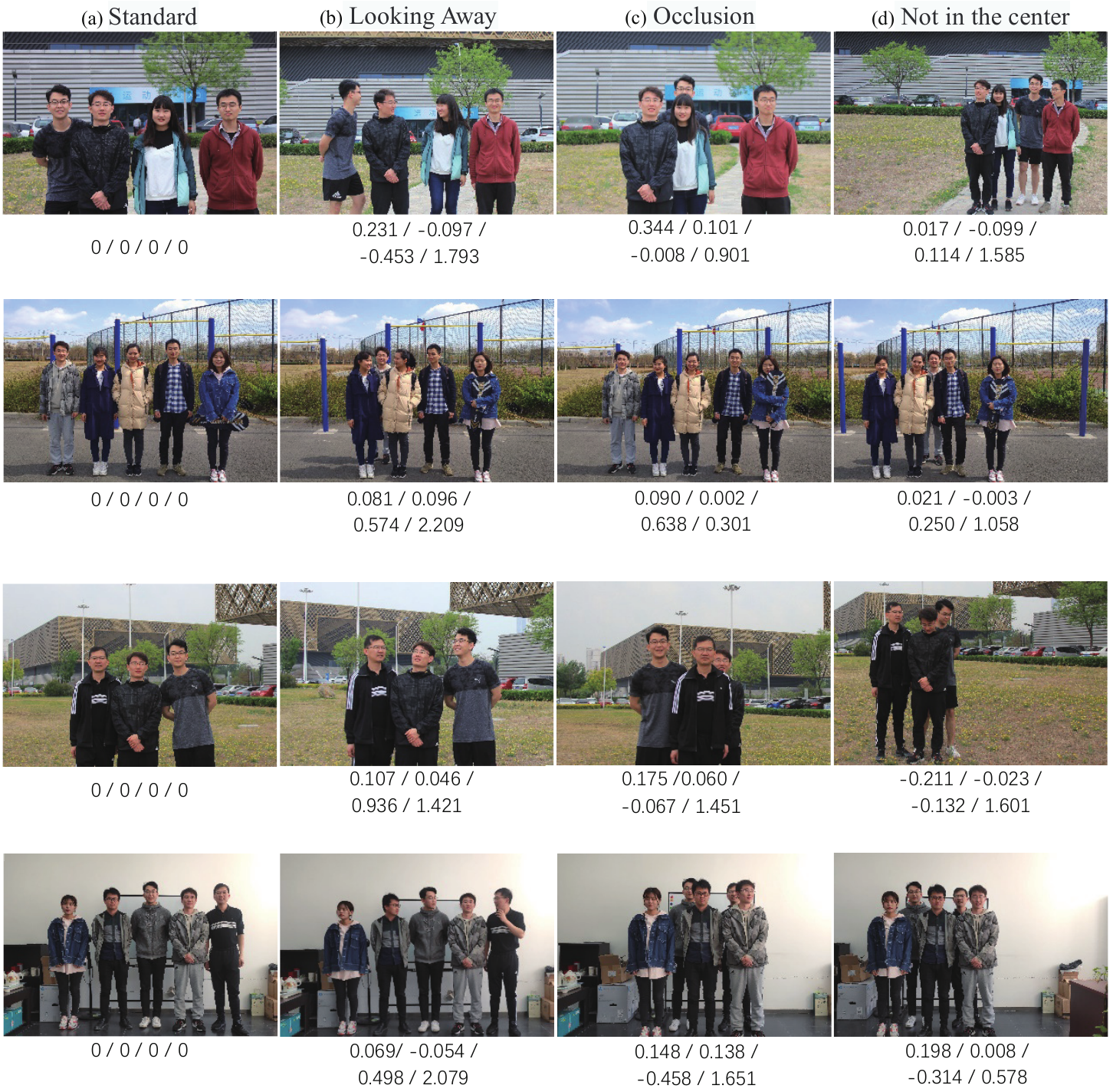}
	\caption{Compared with the results of the deep learning method, the score at the bottom of the picture represents $\delta_{NIMA-RES}$/ $\delta_{NIMA-mobile}$ / $\delta_{Kong}$ / $\delta_{our}$}
	\label{8}
\end{figure}

Looking at Figure 8, taking the first row (a) as an example, the degree of differentiation using the deep learning method are very small or even negative which are 0.231, -0.097, -0.453, respectively. It shows that these methods only from the perspective of the general image to assess the photo, do not consider the people's state. The $\delta$ of our regression model can reach 1.793, which makes a good distinction between standard group photo and “Looking away”. This is mainly because our assessment method is based on the constraint of people's state, then combined with general features to assessment group photos. It can be found from the observation of column (c) that the face in first group, the third group and the fourth group have serious occlusion. Using our method to evaluate, the discrimination is close to 1. In the second group, the rightmost character is slightly obscured by objects, and the discrimination is 0.301. However, the discrimination of the deep learning method in the evaluation of such photos is small, all of which are floating up and down 0, and there is Irregular, which proved that the occlusion feature is also effective in group photo evaluation. From Figure 8, it can be seen that the discrimination (in the range of [1.4-2.3]) of column (a) is generally higher than that of column (b) and (c) (in the range of [0.3-1.7]). This fully corresponds our expectations, as well as the importance ranking of photo features, the impact of eye`s state is greater than the face occlusion and the position of the person on the photo assessment. We also observed that when assess column (d), there's a good chance that $\delta$ being negative , which indicates that the deep learning methods consider that the object on the side has a higher aesthetic score than the object on the center. The rule of thirds may be effective when assessment landscape photos, but it is not applicable in the group photos. It also demonstrated that the method based on deep learning relies on a large number of aesthetic photos, without professional knowledge, so it only learns some generic shooting rules and aesthetic features, and it is difficult to make a correct assessment of images in a specific field. On the whole, the discrimination of the assessment method based on deep learning is between -0.5 and 1. The assessment of group photos does not take into account the state of people, and can't distinguish between good and bad photos when assess multiple group photos in the same scene, but the $\delta$ of our model is between 0.3 and 2.3, which can make a good discrimination of such photos.

\section{Conclusion and future work}
In this work, by analyzing the aesthetic features of group photography, we address the problem that the general method cannot accurately evaluate the group photograph, and introduce group photography features to facilitate investigation of this problem. Furthermore, we construct a group photography dataset (GPD), and built an online annotation tool for collecting the label of GPD. In the experiments, we validated that the proposed method can better evaluate group photography than previous methods that only considered generic features. However, our group photography scene is relatively single. Moreover, there is still a lot of space for improvement in the extraction of group photography features and generic aesthetic  features in the future. To further improve the accuracy of the aesthetic evaluation of group photography, much work remains to be done.
\section*{Acknowledgments}
This research was partially supported by National Natural Science Foundation of China [Grant No. 61771340,61602344] and Natural Science Foundation of Tianjin, China [Grant No. 18JCYBJC15300].

\bibliography{mybibfile}

\end{document}